\documentclass[a4paper]{llncs}

\usepackage{siunitx}
\usepackage{graphicx}
\usepackage{tabulary}
\usepackage[noend]{algpseudocode}
\usepackage{algorithm}
\usepackage{amsmath}
\usepackage{subfigure}
\usepackage{url}

\title{\LARGE \bf
Deep Head Pose Estimation \\ from Depth Data \\ for In-car Automotive Applications
}

\author{Marco Venturelli, Guido Borghi, Roberto Vezzani, Rita Cucchiara}
\institute{DIEF - University of Modena and Reggio Emilia \\ 
Via P. Vivarelli 10, 41125 Modena, Italy \\ 
Email: \{name.surname\}@unimore.it}

\begin{document}

\maketitle
\thispagestyle{empty}
\pagestyle{empty}

\begin{abstract}

Recently, deep learning approaches have achieved promising results in various fields of computer vision. In this paper, we tackle the problem of head pose estimation through a Convolutional Neural Network (CNN). Differently from other proposals in the literature, the described system is able to work directly and based only on raw depth data. Moreover, the head pose estimation is solved as a regression problem and does not rely on visual facial features like facial landmarks. We tested our system on a well known public dataset, \textit{Biwi Kinect Head Pose}, showing that our approach achieves state-of-art results and is able to meet real time performance requirements.  
\end{abstract}

\section{INTRODUCTION}
\begin{figure}[bth]
\centering
\subfigure[]{\includegraphics[width=0.3\columnwidth]{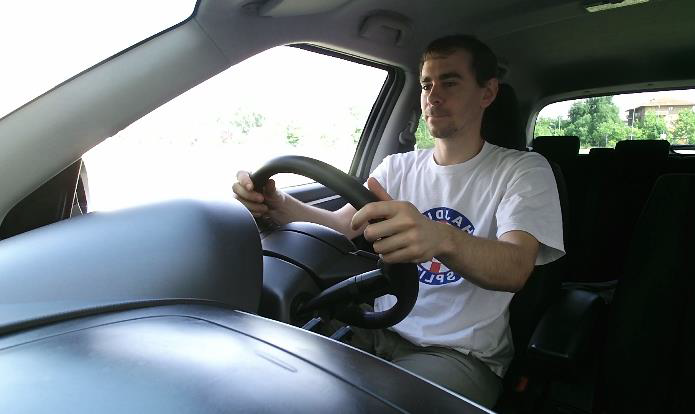}} 
\subfigure[]{\includegraphics[width=0.3\columnwidth]{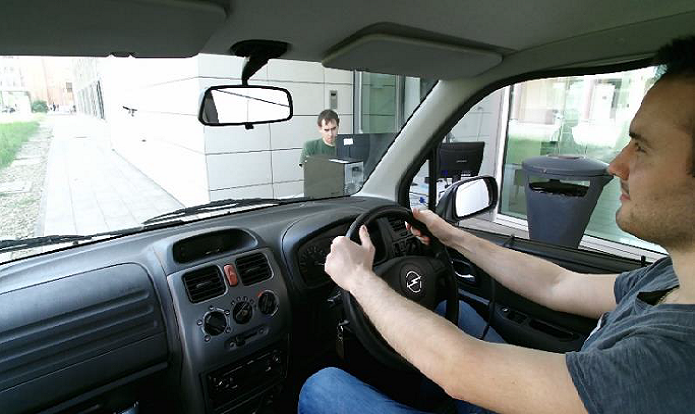}}
\subfigure[]{\includegraphics[width=0.3\columnwidth]{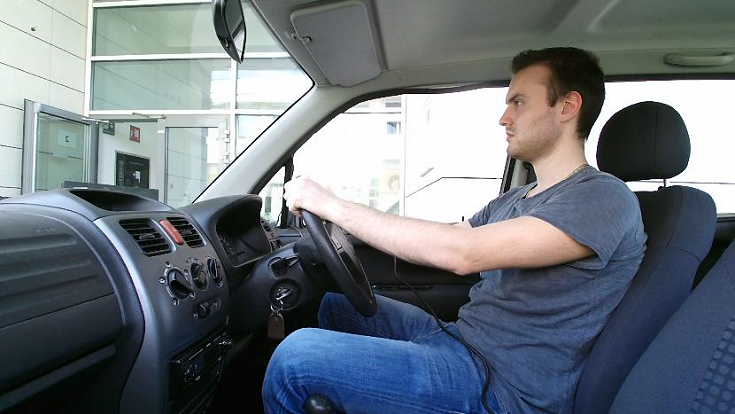}}
\subfigure[]{\includegraphics[width=0.3\columnwidth]{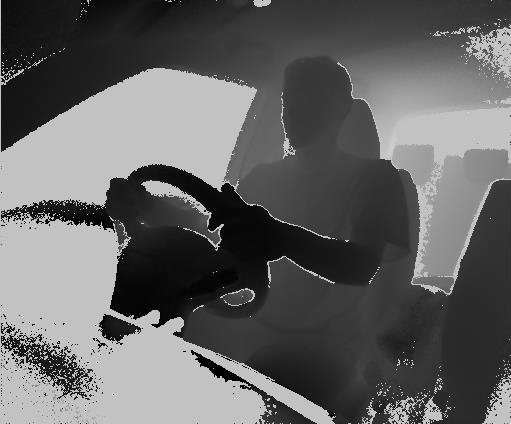}}
\subfigure[]{\includegraphics[width=0.3\columnwidth]{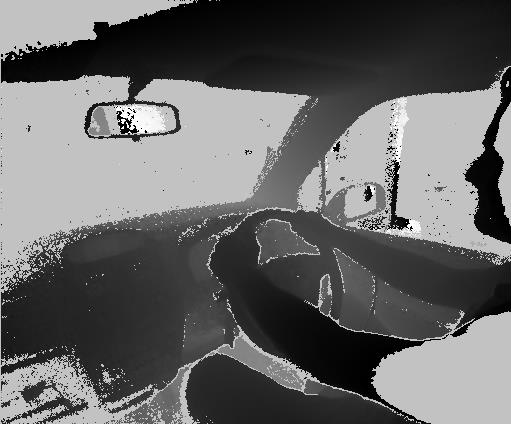}} 
\subfigure[]{\includegraphics[width=0.3\columnwidth]{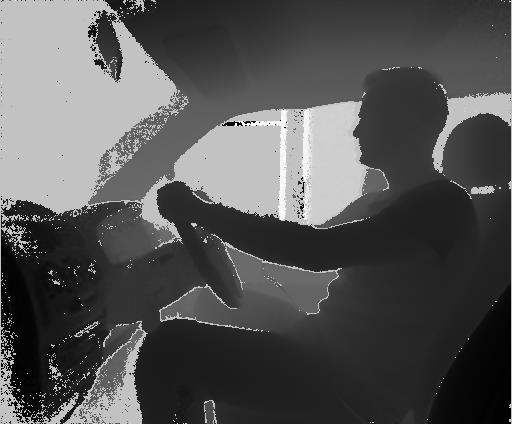}}
\caption{Images acquired with \textit{Microsoft Kinect One} device from different in-cockpit points of view. In the first row, RGB images are reported, while in the second row the corresponding depth maps are shown. It can be noted how light and the position of the camera could influence the view quality of the head and the other visible body parts and produce partial or severe occlusions.  } 
\label{fig:EcUND} 
\end{figure}

Head pose estimation is an important visual cue in many fields, such as human intention, motivation, attention and so on. In particular, in automotive context, head pose estimation is one of the key elements for attention monitoring and driver behavior analysis.\\ Distracting driving has a crucial role in road crashes, as reported by the official US government website about distracted driving \cite{DD}. In particular, 18\% of injury crashes were caused by distraction, more than 3000 people were killed in 2011 in a crash involving a distracted driver, and distraction is responsible for 11\% of fatal crashes of drivers under the age of twenty \cite{DD}. The \textit{National Safety Administration} (NHTSA) defines driving distraction as "\textit{an activity that could divert a person's attention away from the primary task of driving}".\\
Driving distractions have been classified into three main categories \cite{CrayeK15}:
\begin{itemize}
\item \textbf{Manual Distraction}: the hands of the driver are not on the wheel;  examples of this kind of activity are text messaging or incorrect use of infotainment system (radio, GPS navigation device and others).
\item \textbf{Visual Distraction}: the driver does not look at the road, but, for example, at the smart-phone screen or a newspaper.
\item \textbf{Cognitive Distraction}: the driver is not focused on driving activity; this could occur if talking with passengers or due to bad physical conditions (torpor, stress).
\end{itemize}
It is intuitive that smartphone is one of the most important cause of fatal driving distraction: it involves all three distraction categories mentioned above and it represents about 18\% of fatal driver accidents in North America.\\
The introduction of semi-autonomous and autonomous driving vehicles and their coexistence with traditional cars is going to increase the already high interest about driver attention studies. Very likely, automatic pilots and human drivers will share the control of the vehicles, and the first will need to call back the latter when needed. For example, the same situation is currently happening on airplanes. The monitoring of the driver attention level is a key-enabling factor in this case. In addition, legal implications will be raised\cite{Rahman4021}.\\
Among the others, a correct estimation of the driver head pose is an important element to accomplish driver attention and behavior monitoring during driving activity. To this aim, the placement and the choice of the most suitable sensing device is crucial. In particular, the final system should be able to work on each weather condition, like shining sun and clouds, in addition to sunrises, sunsets, nights that could dramatically change the quality and the visual performance of acquired images. Infrared or, even better, depth cameras overcome classical RGB sensors in this respect (see Fig. \ref{fig:EcUND}). \\ 

In this work, we use and investigate the potentiality of depth images. The release of cheap but accurate 3D sensors, like \textit{Microsoft Kinect}, brings up new opportunities in this field and much more accurate depth maps. Existing depth-based methods either need manual or semi-automatic initialization, could not handle large pose variations and does not work in real time; all these elements are not admissible in automotive context. \\
Here, we propose an efficient and accurate head localization framework, exploiting Convolutional Neural Networks (CNN) in a data regression manner. The provided results confirm that a low quality depth input image is enough to achieve good performance. Although the recent advantages in classification tasks using CNNs, the lack of research on deep approaches for angle regression proves the complexity of this kind of task. \\

\section{RELATED WORK}
Head localization and pose estimation are the goal of several works in the literature \cite{Trivedi2009}. Existing methods can be divided depending on the type of data they rely on: 2D (RGB or gray scale), 3D (RGB-D) data, or both. Due to the approach of our work, we briefly describe methods that use 2D data, and we focus on methods based on depth data or a combination of depth and intensity information.
In general, methods relying solely on RGB images are sensitive to illumination, lack of features and partial occlusions \cite{fanelli2011}. \\
To avoid these issues, \cite{ahn2014} use for the first time Convolutional Neural Network to exploit CNN well-known power in space and color invariance. This is one of the first case in which a Convolutional Neural Network (CNN) is used in order to perform head pose estimation using images acquired by a monocular camera. This architecture is exploited in a data regression manner. A mapping function between three head predicted angles and visual appearance is learned. Despite the use of deep learning techniques, system is working in real time with the aid of a GPU. Also in \cite{mukherjee2015} a CNN is exploited to predict head pose and gaze direction, a regression technique is approximated with a Softmax layer with 360 classes. \\
Besides, CNN is used in \cite{liu3d2016}: the network is trained on synthetic images. Recently, the use of synthetic dataset is increasing to support deep learning approaches that basically require huge amount of data. 
In \cite{chen2016} the problem of head pose estimation is taken on extremely low resolution images, achieving results very close to state-of-art results for full resolution images.
\cite{drouard2015} used HOG features and a Gaussian locally-linear mapping model, learned using training data, to map the face descriptor onto the space of head poses and to predict angles of head rotation.
\\
Malassiotis et al. in \cite{malassiotis2005} proposed a method based on low quality depth data to perform head localization and pose estimation; this method relies on the accurate localization of the nose and that could be a strong limitation for automotive context.
Bretenstein et al. in \cite{breitenstein2008} proposed a real time method which can handle large pose variation, partial occlusion and facial expression from range images; the main issue is that the nose must to be always visible, this method uses geometric features to generate nose candidates which suggest head position hypothesis. The alignment error computation is demanded to a dedicated GPU in order to work in real time.\\
\cite{kondori2011} investigated an algorithm based on least-square minimization of the difference between the measured rate of change of depth at a point and the rate predicted, to perform head localization and then head detection and tracking during videos. 
Fanelli et al. in \cite{fanelli2011} proposed a real time framework based on Random Regression Forests to perform head pose estimation from depth images.\\
In \cite{padeleris2012} the head pose estimation is treated as a optimization problem that is solved through Particle Swarm Optimization. This method needs a a frame (the first of the sequence) to construct the reference head pose from depth data; low real time performance are obtained thanks to a GPU.
Papazov et al. in \cite{papazov2015} introduced a novel triangular surface patch descriptor to encode shapes of 3D surface; this descriptor of an input depth map is matched to the most similar ones that were computed from synthetic head models in a training phase.
\\
Seeman et al. in \cite{Seemann2004} proposed a method based on neural network and a combination of depth information, acquired by a stereo camera, and skin color histograms derived from RGB images; in this case work limit is that the user face has to be detected in frontal pose at the beginning of framework pipeline.
\cite{bleiweiss2010} presented a solution for real time head pose estimation based on the fusion of color and time-of-flight depth data. The computation work is demanded to a dedicated GPU.\\
Baltrusaitis et al. in \cite{baltruvsaitis2012} presented a 3D constrained local method for robust facial feature tracking under varying poses, based on the integration both depth and intensity information. In this case the head pose estimation is one of the consequences of landmark tracking.
A method to elaborate HOG features both on 2D (intensity) and depth data is described in \cite{yang2012,saeed2015}; in the first case a Multi Layer Perceptron is the used for classification task; in the second, a SVM is used.
Ghiass et al. \cite{ghiass2015} performed pose estimation by fitting a 3D morphable model which included pose parameter, starting both from RGB and depth data. This method relies on detector of Viola and Jones \cite{viola2004}.

\section{HEAD POSE ESTIMATION}
The goal of the system is the estimation of the head pose (i.e., pitch, roll, yaw angles with respect to a frontal pose) directly from depth data using a deep learning approach. We suppose to have a correct head detection and localization. The description of these steps are out of the scope of this paper. Differently from \cite{Seemann2004,malassiotis2005,breitenstein2008}, additional information such as facial landmarks, nose tip position, skin color and so on are not taken into account. 

\subsection{Image pre-processing} \label{sec:pose_prepro}
\begin{figure}[bh!]
\centering
\subfigure[]{\includegraphics[width=0.15\columnwidth]{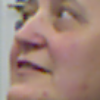}} 
\subfigure[]{\includegraphics[width=0.15\columnwidth]{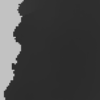}}
\subfigure[]{\includegraphics[width=0.15\columnwidth]{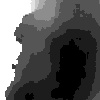}}
\subfigure[]{\includegraphics[width=0.15\columnwidth]{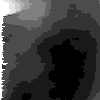}}\\
\subfigure[]{\includegraphics[width=0.15\columnwidth]{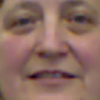}} 
\subfigure[]{\includegraphics[width=0.15\columnwidth]{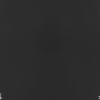}}
\subfigure[]{\includegraphics[width=0.15\columnwidth]{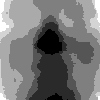}}
\subfigure[]{\includegraphics[width=0.15\columnwidth]{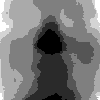}}
\caption{Example of image pre-processing on two different input cropped image: (a) is the RGB frame, (b) the correspondent depth map, (c) depth map after normalization and (d) depth map after the linear interpolation. From (e) to (h) is the same, but with a frontal point of view. It can be noted that in (h) interpolation does not change a lot the visual result, due to the absence of background.} 
\label{fig:facciotti} 
\end{figure}

Image pre-processing is a fundamental step to obtain better performance with the further exploitation of CNN \cite{krizhevsky2012}. \\
First of all, the face images are cropped using a dynamic window.  
Given the center $x_c, y_c$ of the face, the image is cropped to a rectangular box centered in $x_c, y_c$ with width and height computed as: 
$$w, h = \frac{f_{x,y}R}{Z},$$
where $f_{x,y}$ are the horizontal and vertical focal lengths (in pixels) of the acquisition device, $R$ is the width of a generic face (120 mm in our experiments, \cite{ahn2014}) and $Z$ is the distance between the acquisition device and the user obtained from the depth image. The output is an image which contains very few parts of background. Then, the cropped images are resized to 64x64 pixels and their values are normalized so that the mean and the variance are 0 and 1, respectively. This normalization is also required by the specific activation function of the network layers (see Section \ref{sec_arch}). 
Finally, to further reduce the impact of the background pixels, each image row is linearly stretched (see Algorithm \ref{al:linearstretch}) keeping only foreground pixels. Some example results are reported in Figure \ref{fig:facciotti}.

\begin{algorithm}
\caption{Linear Interpolation Algorithm}
\begin{algorithmic}[1]
\Procedure{linear interpolation}{}
\State $w$ : image width
\For{$row$ in image rows}
\State $x_{min}$ = first foreground pixel in $row$
\State $x_{max}$ = last foreground pixel in $row$
\For{$x$=0 to w-1}
\State $x_{src} = x/ (w-1)*(x_{max}-x_{min})$
\State $x_1 = \lfloor x_{src} \rfloor$
\State $x_2 = x_1 + 1$
\If{$x_2 \leq w$}
	\State $\lambda = x_2 - x_{src}$
	\State $row_{out}[x] = row[x_1]* \lambda +row[x_2]*(1-\lambda)$
\Else
	\State $row_{out}[x] = row[x_1]$
\EndIf
\EndFor
\EndFor

\EndProcedure
\end{algorithmic}
\label{al:linearstretch}
\end{algorithm}

\subsection{Deep Architecture} \label{sec_arch}
\begin{figure}[h!]
    \centering
    \includegraphics[width=1\textwidth]{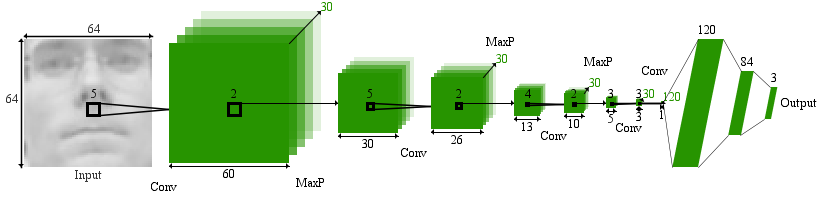}
    \caption{The deep architecture that represents network adopted in our work: input is a 64x64 image, there are 5 convolutional layers, 3 fully connected layers; output has 3 neurons to predict head yaw, pitch and roll. This chart is obtained with \textit{DeepVisualizer}.}
    \label{fig:rete}
\end{figure}

The architecture of the neural network is inspired from the one proposed by Ahn \textit{et al.} \cite{ahn2014}. We adopt a shallow deep architecture, in order to obtain a real time system and to maintain good accuracy. The network takes images of 64x64 pixels as input, which are relatively smaller than other deep architecture for face applications. \\
The proposed structure is depicted in Figure \ref{fig:rete}.  It is composed of 5 convolutional layers; the first four have 30 filters whereas the last one has 120 filters. At the end of the network there are three fully connected layers, with 120, 84 and 3 neurons respectively, that correspond to the three head angles (yaw, pitch and roll). The size of the convolution filters are 5x5, 4x4, 3x3, depending on the layer. Max-pooling is conducted only three times. The activation function is the hyperbolic tangent: in this way network can map output $[-\infty, +\infty] \rightarrow [-1, +1]$, even if ReLU tends to train faster that other activation functions \cite{krizhevsky2012}. In this way, the network outputs  continuous instead of discrete values. We adopt the Stochastic Gradient Descent (SGD) as in \cite{krizhevsky2012} to resolve back-propagation.\\ 
A L2 loss function is exploited:
$$loss = \sum_i^n \lVert \mathbf{y_i - f(x_i)} \rVert ^ 2 _2.$$

The network representation in Figure \ref{fig:rete} is obtained using \textit{DeepVisualizer}, a software we recently developed. \footnote{ The tool is written in Java and it is completely free and open source.
It takes as input the JSON file produced by the Keras framework and generates image outputs in common formats such as png, jpeg or gif. We invite the readers to test and use this software, hoping it can help in deep learning studies and presentations. The code can be downloaded at the following link: \\
\url{http://imagelab.ing.unimore.it/deepvisualizer}
}

\subsection{Training}
The network has been trained with a batch size of 64, a decay value of $5^{-4}$, a momentum value of $9^{-1}$ and a learning rate set to $10^{-1}$, descending to $10^{-3}$ in the final epochs \cite{krizhevsky2012}. Ground truth angles are normalized to $[-1, +1]$.\\
An important and fundamental aspect of deep learning is the amount of training data. To this aim, we performed data augmentation to avoid over fitting on limited datasets. For each pre-processed input image, 10 additional images are generated. 5 patches are randomly cropped from each corner and from their center, other 4 patches are extracted by cropping original images starting from the bottom, upper, left and right image part. Finally, one more patch is created adding Gaussian noise (\textit{jittering}). 

\begin{figure}[h!]
    \centering
    \includegraphics[width=0.65\columnwidth]{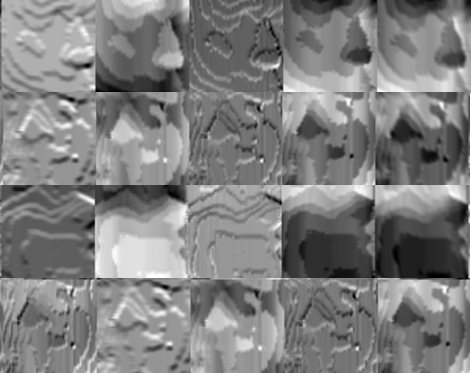}
    \caption{Some example of feature maps generated by our network. The network has learned to extract facial elements like nose tip, eye holes, cheeks and contour lines.}
    \label{fig:featuremap}
\end{figure}

\section{EXPERIMENTAL RESULTS}
In order to evaluate the performance of the presented method, we use a public dataset for head pose estimation that contain both RGB and depth data, namely \textit{Biwi Kinect Head Pose Database}.

\subsection{Biwi Kinect Head Pose Database}
\begin{figure}[t!]
\centering
\subfigure[]{\includegraphics[width=0.3\columnwidth]{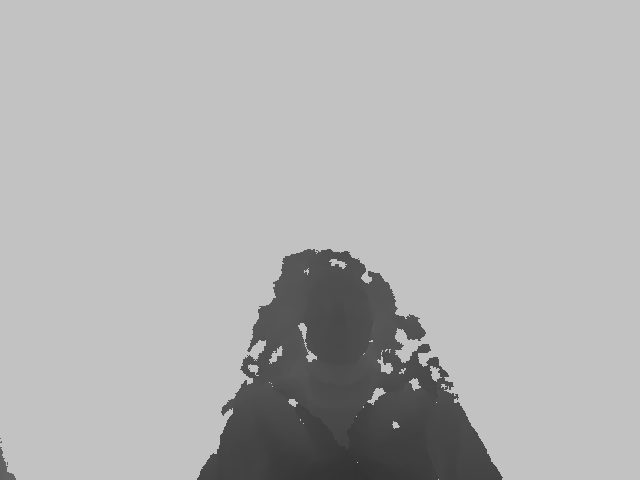}} 
\subfigure[]{\includegraphics[width=0.3\columnwidth]{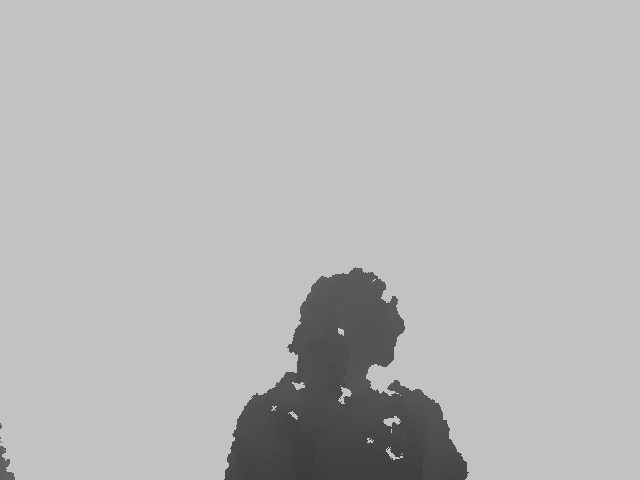}}\\
\subfigure[]{\includegraphics[width=0.3\columnwidth]{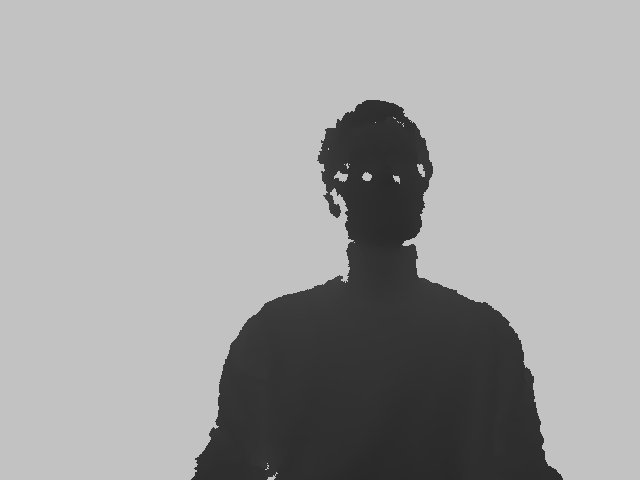}}
\subfigure[]{\includegraphics[width=0.3\columnwidth]{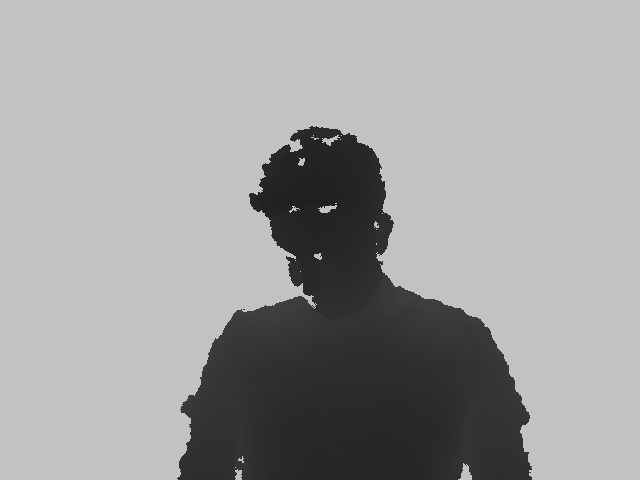}}
\caption{Some example of \textit{Biwi} dataset frames that present visual artifacts, like holes, with female (a - b) and male (c - d) subjects.} 
\label{fig:biwi_errors} 
\end{figure}

This dataset is introduced by Fanelli \textit{et al.} in \cite{fanelli2013} and it is explicitly designed for head pose estimation task. It contains 15678 upper body images of 20 people (14 males and 6 females) and 4 people were recorded twice. The head rotation spans about $\pm \ang{75}$ for yaw, $\pm \ang{60}$ for pitch and $\pm \ang{50}$ for roll. For each frame a depth image and the corresponding RGB image are provided, acquired sitting in front  a stationary \textit{Microsoft Kinect}; both of them have a resolution of 640x480. Besides ground truth pose angles, calibration matrix and head center - the position of the nose tip - are given. \\ 
This is a challenging dataset because of the low quality of depth images (\textit{e.g.} long hair in female subjects cause holes in depth maps, some subject wear glasses, see Figure \ref{fig:biwi_errors}); besides the total number of samples used for training and testing and the subjects selection are not clear, even in the original work \cite{fanelli2011}. To avoid this ambiguity, we use sequences 1 and 12 to test our network, which correspond to not repeated subjects. Some papers use own method to collect results (\textit{e.g.} \cite{ahn2014}), so their results are not reported and compared.

\subsection{Discussion about other datasets}
\begin{figure}[h!]
\centering
\subfigure[]{\includegraphics[width=0.40\columnwidth]{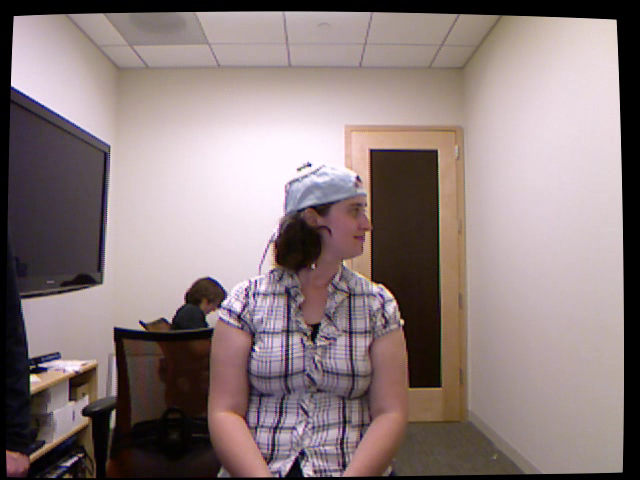}} 
\subfigure[]{\includegraphics[width=0.40\columnwidth]{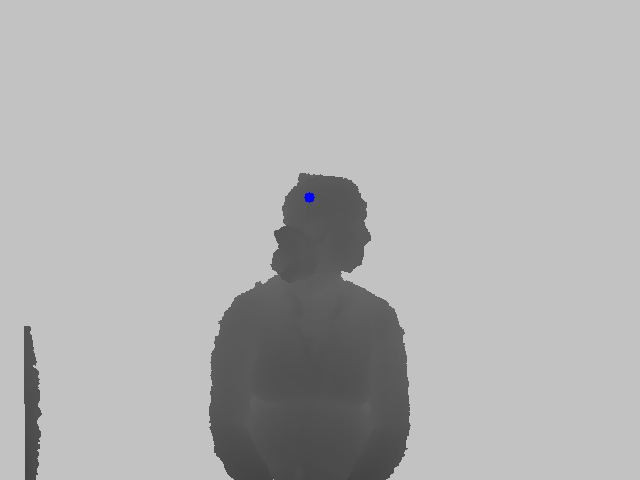}}
\caption{Two frames of the \textit{ICT-3DHP Dataset}. At the right can be seen the white cap, at the left the correspondent head center position that is translated to the left.} 
\label{fig:3dhp} 
\end{figure}

Several dataset for head pose estimation were collected in this decade, but in most cases there are some not desirable issues. The main issues are that they do not provide depth data (\textit{e.g.}, \textit{RobeSafe Driver Monitoring Video Dataset} \cite{Nuevo2010}), or that not all angles are included (\textit{e.g.}, \textit{Florence 2D/3D Face Dataset} \cite{Masi} reports only yaw angles). Moreover, most of the datasets have not enough frames or images for deep learning approaches.\\
\textit{ICT-3DHP Dataset} \cite{baltruvsaitis2012} is collected using \textit{Microsoft Kinect} sensor. It contains about 14000 frames both intensity and depth. The ground truth is labeled using a \textit{Polhemus Fastrack} flock of birds tracker. This dataset has three main drawbacks: users had to wear a white cap for the tracking system. The cap is well visible both in RGB and depth video sequences. Second, there is a lack of training data images with roll angles and the head center position is not so accurate (see Figure \ref{fig:3dhp}). Finally, this dataset is not good for deep learning, because of its small size and the presence of few subjects. 

\subsection{Quantitative results}

\begin{table}[h]
\caption{Results on \textit{Biwi} dataset (Euler angles)}
\centering
\begin{tabular}{|c|c|c|c|c|c|}
\hline
\textbf{Met.} & \textbf{Data} &\textbf{Pitch} & \textbf{Roll}  & \textbf{Yaw}   & \textbf{Time}    \\ \hline
\cite{saeed2015}              & RGB+depth 			& 5.0 $\pm$ 5.8			& 4.3 $\pm$ 4.6			& 3.9 $\pm$ 4.2		& -\\ \hline
\cite{mukherjee2015}          & RGB+depth 				& 4.76					& -			& 5.32		& -\\ \hline
\cite{fanelli2011}            & depth 					& 8.5 $\pm$ 9.9			& 7.9 $\pm$ 8.3			& 8.9 $\pm$ 13.0	& 40 ms/frame  \\ \hline
\cite{yang2012}				  & RGB+depth				& 9.1 $\pm$ 7.4			& 7.4 $\pm$ 4.9			& 8.9 $\pm$ 8.2		& 100 ms/frame \\ \hline
\cite{baltruvsaitis2012}      & RGB+depth				& 5.1 					& 11.2					& 6.29				 & -\\ \hline
\cite{papazov2015}      	  & depth				    & 3.0 $\pm$ 9.6 		& 2.5 $\pm$ 7.4			& 3.8 $\pm$ 16.0 	& 76 ms/frame\\ \hline
Our            				  & depth					& \textbf{2.8} $\pm$ \textbf{3.1}			& \textbf{2.3} $\pm$ \textbf{2.9}			& \textbf{3.6} $\pm$ \textbf{4.1}		& \textbf{10 ms/frame}\\ \hline
\end{tabular}
\label{tab:resBiwiPose}
\end{table}

\begin{figure}[h]
    \centering
    \includegraphics[width=1\textwidth]{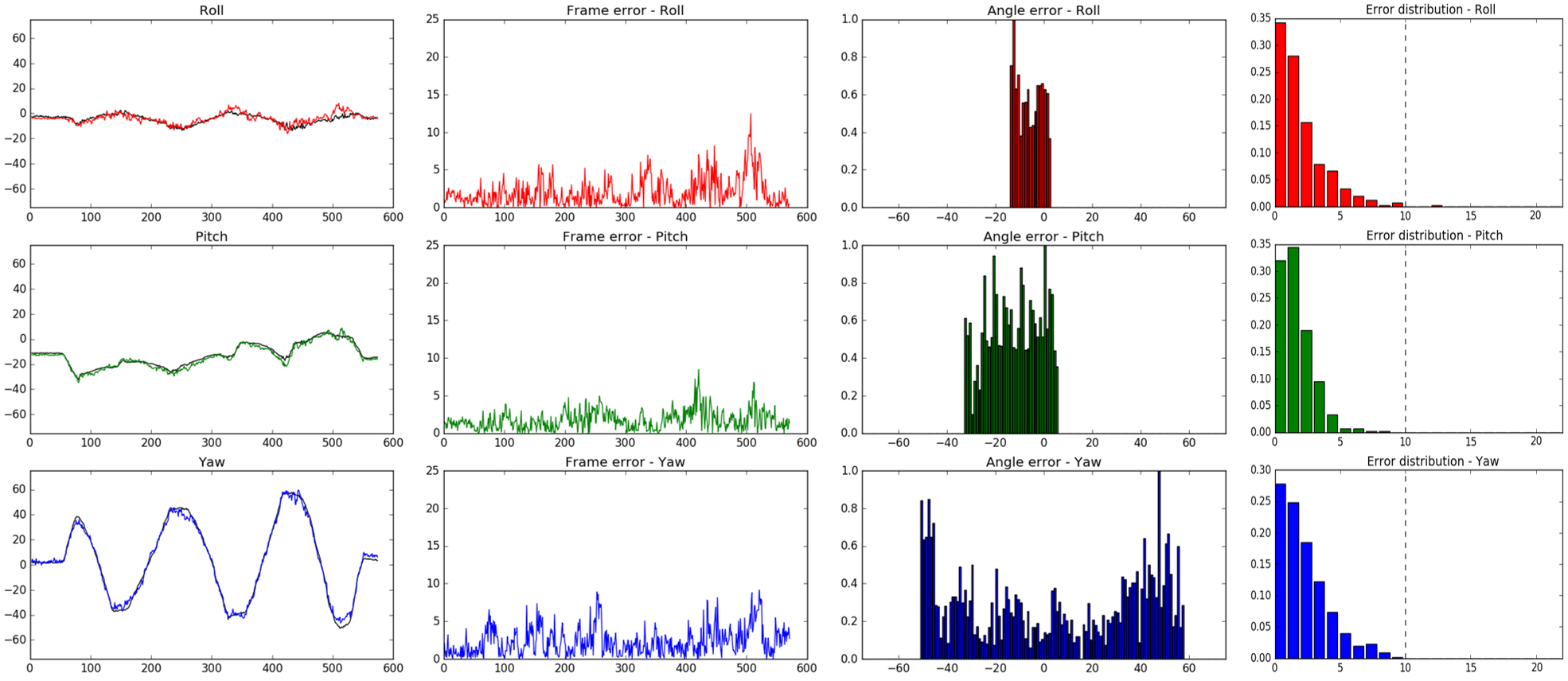}
    \caption{Experimental results: roll, pitch and yaw angles are reported on the three rows. The ground truth is superimposed in black. The angle error per frame is reported in the second column, while in the third column histograms highlights the errors at specific angles. The error distribution is reported in the last column.}
    \label{fig:grafici}
\end{figure}

\begin{figure}[h]
    \centering
    \includegraphics[width=1\textwidth]{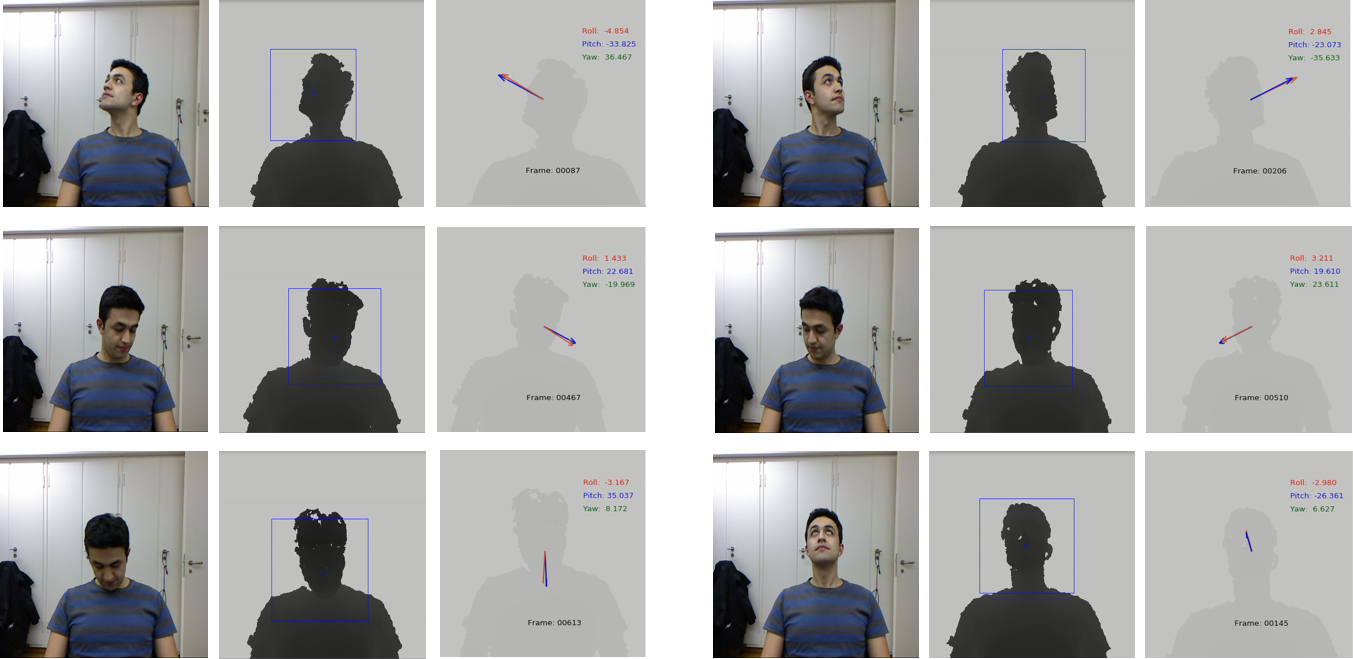}
    \caption{The first column show RGB frames, the second the correspondent depth map frame: blue rectangle reveal the dynamic crop to extract the face. The last column reports yaw (red), pitch (blue) and roll (green) angles values and the frame number (\textit{Biwi} dataset)}
    \label{fig:demo}
\end{figure}

Table \ref{tab:resBiwiPose} reports the results obtained on \textit{Biwi Kinect Head Pose Dataset}. We follow the evaluation protocol proposed in \cite{fanelli2011}. Processing time is tested on \textit{Nvidia Quadro k2200} 4GB GPU with the same test sequences.\\
Results reported in Table \ref{tab:resBiwiPose} show that our method overcomes other state-of-the-art techniques, even those working on both RGB and depth data or are based on deep learning approaches \cite{mukherjee2015}.\\ 
Thanks to the high accuracy reached, the proposed network can be used for efficient and precise head orientations applications, also in automotive context, with an impressively low elaboration time.\\
Figure \ref{fig:demo} shows an example of working framework for head pose estimation in real time: head center is taken thanks to ground truth data; the face is cropped from raw depth map (in the center image, the blue rectangle) and in the right frame yaw, pitch and roll angles are shown.

\section{CONCLUSIONS}
We present a innovative method to directly extract head angles from depth images in real time, exploiting  a deep learning approach. Our technique aim to deal with two main issue of deep architectures in general, and CNNs in particular: the difficulty to solve regression problems and the traditional heavy computational load that compromises real time performance for deep architectures.\\
Our approach is based on Convolutional Neural Network with shallow deep architecture, to preserve time performance, and is designed to resolve a regression task.\\
There is rich possibility for extensions thanks to the flexibility of our approach: in future work we plan to integrate temporal coherence and stabilization in the deep learning architecture, maintaining real time performance, incorporate RGB or infrared data to investigate the possibility to have  a light invariant approach even in particular context (\textit{e.g.} automotive). Head localization through deep approach could be studied in order to develop a complete framework that can detect, localize and estimate head pose inside a cockpit.\\
Besides studies about how occlusions can deprecate our method are being conducted.

\bibliographystyle{IEEEtran}
\bibliography{bibliography}

\end{document}